\documentclass[english]{elsarticle}
\usepackage[T1]{fontenc}
\usepackage[latin9]{inputenc}
\usepackage{array}
\usepackage{float}
\usepackage{multirow}
\usepackage{amsmath}
\usepackage{amssymb}
\usepackage{graphicx}
\usepackage{setspace}
\PassOptionsToPackage{normalem}{ulem}
\usepackage{ulem}

\makeatletter

\AtBeginDocument{\providecommand\figref[1]{\ref{fig:#1}}}
\providecommand{\tabularnewline}{\\}
\floatstyle{ruled}
\newfloat{algorithm}{tbp}{loa}
\providecommand{\algorithmname}{Algorithm}
\floatname{algorithm}{\protect\algorithmname}

\@ifundefined{date}{}{\date{}}
\@ifundefined{showcaptionsetup}{}{%
 \PassOptionsToPackage{caption=false}{subfig}}
\usepackage{subfig}
\makeatother

\usepackage{babel}
\begin{document}
\begin{frontmatter}

\title{A Kalman filtering induced heuristic optimization based partitional
data clustering }

\author[1]{Arjun Pakrashi\corref{corauth}} \ead{arjun.pakrashi@ucdconnect.ie,arjun.pakrashi@insight-centre.org}
\author[2]{Bidyut B. Chaudhuri} \ead{bbc@isical.ac.in}
\address[1]{Insight Centre for Data Analytics, University College Dublin, Ireland}
\address[2]{Computer Vision \& Pattern Recognition Unit, Indian Statistical Institute, 203 B.T. Road, Kolkata 700108, India}
\cortext[corauth]{Corresponding Author}
\begin{abstract}
Clustering algorithms have regained momentum with recent popularity
of data mining and knowledge discovery approaches. To obtain good
clustering in reasonable amount of time, various meta-heuristic approaches
and their hybridization, sometimes with K-Means technique, have been
employed. A Kalman Filtering based heuristic approach called Heuristic
Kalman Algorithm (HKA) has been proposed a few years ago, which may
be used for optimizing an objective function in data/feature space.
In this paper at first HKA is employed in partitional data clustering.
Then an improved approach named HKA-K is proposed, which combines
the benefits of global exploration of HKA and the fast convergence
of K-Means method. Implemented and tested on several datasets from
UCI machine learning repository, the results obtained by HKA-K were
compared with other hybrid meta-heuristic clustering approaches. It
is shown that HKA-K is atleast as good as and often better than the
other compared algorithms.
\end{abstract}
\begin{keyword}
Clustering, K-Means, Optimization, Metaheuristic Optimization, Heuristics
\end{keyword}
\end{frontmatter}

\section{Introduction}

Clustering is the process of assigning a set of $n$ data points into
$C$ classes based on the similarity between the data points in the
feature space. It is useful when some prototype data from known classes
are not available for training a supervised classifier or for an exploratory
data analysis task. It is one of the earliest pattern classification
approaches and has found renewed interest since the beginning of data
mining and big data analytics. 

Clustering can be broadly classified into two categories: \emph{Hierarchical}
and \emph{Partitional}. Hierarchical clustering methods are again
grouped into two types, namely \emph{divisive} or top-down and \emph{agglomerative}
or bottom-up methods. If the number of desired clusters is known (say,
$K$) a priori, the approach can be made non-hierarchical and the
data can be assigned into $K$ clusters using a partitional clustering
algorithm. Else, several clusterings are generated and the best among
them is chosen on the basis of some objective criterion.

One of the classical partitional clustering approaches is the K-Means
algorithm, which is a simple way to discover convex, especially hyper-spherical
clusters. It starts with $K$ seed points in the feature space representing
the initial cluster centers. Then it assigns a data point $p$ to
the cluster whose center is nearest to $p$. When all data are assigned
in such a way, the centers are updated by the mean position of current
data assigned to each cluster. Then a new iteration starts and the
data are redistributed using the nearness criterion stated above.
The process is repeated till the centroids do not change anymore or
a specified number of iterations is reached. 

However, K-Means approach may fail to find desired results for some
choice of initial seed points. In such a case it converges to a local
sub-optimal solution, or may even fail to converge \cite{[22]}. One
of the various techniques proposed to tackle the problem model the
clustering task as a non-convex optimization problem employing a cluster
quality based objective function and then try to find its global optimum.
Since finding such an optimum is an NP-Complete problem \cite{[39]},
metaheuristic algorithms are employed, among others, to find a near
optimal solution. A metaheuristic algorithm only uses values of the
objective function to be optimised and does not require the function
to have derivatives. Hence, metaheuristic algorithms are one of the
choices to optimize the non-convex optimization problems. Examples
of metaheuristic based methods are PSO (Particle Swarm Optimization),
GA (Genetic Algorithm), ACO (Ant Colony Optimization) etc. Another
class of algorithms are based on hybrid approach, where two or more
metaheuristic algorithms are combined, or metaheuristic algorithms
are combined with K-Means or other local search methods. Some examples
of this category are hybrid of K-Means and PSO, hybrid of K-Means
and GA as well as PSO-ACO-K-Means hybrid algorithms. A review on different
clustering techniques can be found in \cite{[82]}, while a survey
of metaheuristic algorithms is given in \cite{[55]}. A brief survey
of metaheuristic based clustering algorithms will be presented in
Section \ref{sec:Literature-Survey} of this paper.

A new metaheuristic based optimization approach named Heuristic Kalman
Algorithm (HKA) has been proposed by Toscano and Lyonnet in \cite{[16]}.
This method uses a random number generator with Gaussian probability
distribution to choose a set of points in the search space. At each
subsequent iteration, the mean vector and covariance matrix for the
Gaussian random number generator are modified using a Kalman filtering
framework, so that the mean gets nearer to the global optimum.

In the past, HKA has been employed in applied problems like tuning
of PID controllers \cite{[16]} and in robust observer for rotor flux
estimation of induction machine \cite{[98]}. The present paper uses
HKA for data clustering context. More specifically, the contributions
of this paper are as follows.
\begin{itemize}
\item To employ HKA for data clustering and show that it performs well on
synthetic and real datasets.
\item To propose an improvement over the above HKA based approach through
hybridization with a K-Means like approach. The proposed approach
is named HKA-K where the K indicates application of K-Means like approach.
It is demonstrated that HKA-K converges faster than HKA.
\item To demonstrate that the performance of HKA-K in terms of accuracy,
convergence and computational efficiency is as good as or better than
several other state of the art metaheuristic based clustering algorithms.
\end{itemize}
The rest of the paper is organized as follows. We briefly survey the
metaheuristic algorithms and hybrid metaheuristic algorithms for data
clustering in Section \ref{sec:Literature-Survey}. In Section \ref{sec:HKA}
the HKA based optimization approach is explained by following the
paper \cite{[16]}. A brief exposition of partitional clustering,
along with centroid based K-Means approach is provided in Section
\ref{sec:Clustering-using-HKA}. The approach of using HKA to data
clustering is introduced in the same section. Then the proposed improvement
on HKA clustering, namely HKA-K approach is developed in Section \ref{sec:KHKA-Clustering}.
Experimental results are presented and discussed in Section \ref{sec:Experiment}
where comparisons are also made with other metaheuristic based methods.
Section \ref{sec:Conclusion} contains some concluding remarks on
the paper.

\textbf{}

\section{Brief survey on metaheuristic based clustering\label{sec:Literature-Survey}}

Metaheuristic population based optimization algorithms in general
are used in clustering either as is, or as a hybrid variant combining
multiple algorithms. In general two or more population based algorithm
and/or a local search algorithm are combined together.

Among pure and hybrid metaheuristic based clustering approaches, a
Genetic Algorithm (GA) based approach named COWCLUS \cite{[3]} was
presented, which uses Calinski and Harabasz Variance Ratio Criteria
\cite{[4]} as the cost function and performs standard GA to optimize
it. Another approach in \cite{[31]} has used GA to optimize the clustering
objective function. Yet another GA based approach named GKA \cite{[1]}
has introduced a new mutation operator and combined it with K-Means
for better convergence. These early methods consider each chromosome
representing a clustering of the data points. The chromosome has $n$
alleles, each taking an integer value corresponding to one particular
cluster. On the other hand, the GA based method KGA \cite{[2]} encodes
only the cluster centers in the chromosomes as real numbers. In this
case if $K$ clusters are to be obtained on $d$-dimensional data,
each chromosome will have a length of $(K.d)$. This method combines
K-Means and GA and is more effective than the previous ones.  A Differential
Evolution (DE) and K-Means based algorithm was proposed in \cite{[54]},
where the K-Means method is used to refine the results given by the
recombination operators of DE. 

Particle Swarm Optimization (PSO) has also been employed for data
clustering \cite{[50],[51],[52]}. PSO was hybridized with K-Means
in \cite{[9],[10],[20],[50]} to improve convergence. In \cite{[13]}
a Gaussian based distance metric \cite{[14]} was used while clustering
the data using PSO algorithm. Also, an approach combining GA and PSO
is given in \cite{[29]}. Another hybrid algorithm \cite{[15]} employs
GA, PSO and K-Means approach to determine the number of clusters as
well as to perform clustering.

Among other metaheuristic approaches, Artificial Bee Colony (ABC)
optimization \cite{[40],[41]} was applied for clustering data \cite{[32],[49]}.
Also, an Ant Colony Optimization (ACO) based technique was proposed
in \cite{[42]}. In \cite{[30]} an approach involving PSO, ACO and
K-Means while in \cite{[53]} an ACO and Simulated Annealing (SA)
based algorithm was proposed. The swarm based approach with honey-bee
mating process named Honey Bee Mating Optimization (HBMO) \cite{[43]}
was used in clustering domain \cite{[44]}. 

Clustering was done with Artificial Fish Swarm (AFS) algorithm \cite{[56],[57]}
and Bacteria Foraging Algorithm (BFO) \cite{[58]} as well. Clustering
using Cuckoo Search approach was also proposed in \cite{[61]}. Moreover,
an automatic clustering using Invasive Weed Optimization (IWO) can
be found in \cite{[62]}. A new method, Gravitational Search Algorithm
(GSA), proposed in \cite{[23]}, was used to cluster data \cite{[24]}.
Another approach called Black-Hole search, was also applied in data
clustering \cite{[25]}. 

In the following section we briefly introduce HKA and then explain
how it can be used for clustering.

\section{Heuristic Kalman Algorithm (HKA)\label{sec:HKA}}

The HKA \cite{[16]} is a non-convex, population-based, metaheuristic
optimization algorithm, which works in a manner different from other
population based metaheuristic approaches. Let an objective function
$J$ be defined over a multidimensional and bounded real space $S$.
We have to find the position of a point in $S$ where the value of
$J$ is optimum. The HKA based approach in \cite{[16]} for finding
this point is described below.

A solution is generated through a Gaussian probability density function
(pdf), where the mean vector represents the position in $S$ and the
variance matrix represents the uncertainty of the solution. Given
a search space $S$, the objective of the algorithm is to modify the
mean $\boldsymbol{m}^{(k)}$ and variance matrix $\boldsymbol{P}^{(k)}$
of a Gaussian pdf at $k^{th}$ iteration such that the mean ideally
approaches the global optima within $S$ with respect to the objective
function $J$, along with decreasing variance. The modification of
the mean and variance matrix is done by an approach based on Kalman
filtering. The algorithm starts with an initial mean vector $\boldsymbol{m}^{(0)}$
and variance matrix $\boldsymbol{P}^{(0)}$. The computation of the
initial values are explained later. At each $k^{th}$ iteration the
\emph{measurement} \emph{step} generates $N$ number of points given
by vectors $\boldsymbol{x}(k)=\{\boldsymbol{x}_{1}^{(k)},\boldsymbol{x}_{2}^{(k)},\ldots,\boldsymbol{x}_{N}^{(k)}\}$
from a Gaussian pdf $\mathcal{N}(\boldsymbol{m}^{(k)},\boldsymbol{P}^{(k)})$,
which are ordered such that $J(\boldsymbol{x}_{1}^{(k)})\le J(\boldsymbol{x}_{1}^{(k)})\le\ldots\le J(\boldsymbol{x}_{N}^{(k)})$
is satisfied. Another Gaussian pdf $\mathcal{N}(\boldsymbol{\xi}^{(k)},\boldsymbol{V}^{(k)})$
is generated by taking the mean $\boldsymbol{\xi}^{(k)}$ (Eq. (\ref{eq:measurement}))
of the vector positions of top $N_{\xi}$ number of points, and the
variance $\boldsymbol{V}^{(k)}$ (Eq. (\ref{eq:measurement_error}))
is computed on these top $N_{\xi}$ points. This is considered to
be the \emph{measurement}. Next, in the \emph{estimation step,} the
Kalman filter is used to combine these two Gaussian pdfs to generate
a new Gaussian pdf $\mathcal{N}(\boldsymbol{m}^{(k+1)},\boldsymbol{P}^{(k+1)})$,
where $\boldsymbol{m}^{(k+1)}$ is expected to be closer to the desired
solution than $\boldsymbol{m}^{(k)}$. Here $\boldsymbol{m}^{(k+1)}$
is computed using Eq. (\ref{eq:predicted_mean}), (\ref{eq:m_plus_one})
and (\ref{eq:kalman_gain}) while $\boldsymbol{P}^{(k+1)}$ is computed
using Eq. (\ref{eq:kalman_gain}), (\ref{eq:posterior_variance}),
(\ref{eq:posterior_var}) and (\ref{eq:slowdown}). If a stopping
condition is encountered, then it takes the so far found best value
of $\boldsymbol{m}^{(k)}$, else it carries the Gaussian pdf computed
for the estimation step in the next iteration and repeats the process.

It should be mentioned that, in the current formulation, the matrices
$\boldsymbol{V}^{(k)}$ and $\boldsymbol{P}^{(k)}$ contain only the
diagonal elements and hence can also be treated as vector of dimension
$d$, which is the dimension of the feature space $S$.

The value of $\boldsymbol{\xi}^{(k)}$ is calculated as 

\begin{equation}
\boldsymbol{\xi}^{(k)}=\frac{1}{N_{\xi}}\sum_{i=1}^{N_{\xi}}\boldsymbol{x}_{i}^{(k)}\label{eq:measurement}
\end{equation}

Also, the uncertainty matrix $\boldsymbol{V}^{(k)}$ is given by

\begin{equation}
\boldsymbol{V}^{(k)=}\frac{1}{N_{\xi}}\times\left[\begin{matrix}\sum_{i=1}^{N_{\xi}}(x_{i,1}^{(k)}-\xi_{1}^{(k)})^{2} &  & \mathbf{0}\\
 & \ddots\\
\mathbf{0} &  & \sum_{i=1}^{N_{\xi}}(x_{i,d}^{(k)}-\xi_{d}^{(k)})^{2}
\end{matrix}\right]\label{eq:measurement_error}
\end{equation}

The estimation step computes the parameters $\boldsymbol{m}^{(k+1)}$,
$\boldsymbol{P}^{(k+1)}$ for Gaussian pdf for next iteration by 

\begin{equation}
\hat{\boldsymbol{m}}^{(k)}=\boldsymbol{m}^{(k)}+\boldsymbol{L}^{(k)}(\boldsymbol{\xi}^{(k)}-\boldsymbol{m}^{(k)})\label{eq:predicted_mean}
\end{equation}

\begin{equation}
\boldsymbol{m}^{(k+1)}=\hat{\boldsymbol{m}}^{(k)}\label{eq:m_plus_one}
\end{equation}

\begin{equation}
\boldsymbol{L}^{(k)}=\boldsymbol{P}^{(k)}(\boldsymbol{P}^{(k)}+\boldsymbol{V}^{(k)})^{-1}\label{eq:kalman_gain}
\end{equation}

\begin{equation}
\hat{\boldsymbol{P}}^{(k)}=(\boldsymbol{I}-\boldsymbol{L}^{(k)})\boldsymbol{P}^{(k)}\label{eq:posterior_variance}
\end{equation}

\begin{equation}
\boldsymbol{P}^{(k+1)}=\left(\sqrt{\boldsymbol{P}^{(k)}}+a^{(k)}(\sqrt{\hat{\boldsymbol{P}}^{(k)}}-\sqrt{\boldsymbol{P}^{(k)}})\right)^{2}\label{eq:posterior_var}
\end{equation}

Here $\hat{\boldsymbol{P}}^{(k)}$ is the predicted covariance matrix
and $\boldsymbol{L}^{(k)}$ is the Kalman gain. The details of Kalman
filter and Kalman gain can be found in \cite{[91],[101]}. A \emph{slowdown
factor} $a^{(k)}$ is introduced here to control the fast decrease
of the covariance matrix, which otherwise could converge prematurely
to a bad solution. The factor is defined as 

\begin{equation}
a^{(k)}=\frac{\alpha\cdot min(1,(\frac{1}{d}\sum_{i=1}^{d}\sqrt{v_{i,i}^{(k)}})^{2})}{min(1,(\frac{1}{d}\sum_{i=1}^{d}\sqrt{v_{i,i}^{(k)}})^{2})+max_{1\le i\le d}(\sqrt{\hat{p}_{i,i}^{(k)}})}\label{eq:slowdown}
\end{equation}

In Eq (\ref{eq:slowdown}), $\alpha$ is the \emph{slowdown co-efficient},
which is a parameter of the algorithm, and $v_{i,i}^{(k)}$ and $\hat{p}_{i,i}^{(k)}$
are the $i^{th}$ diagonal of the matrices $\boldsymbol{V}^{(k)}$
and $\hat{\boldsymbol{P}}^{(k)}$ respectively.

The process is repeated till the value of $k$ reaches a predefined
number, or the best $N_{\xi}$ solutions generated at some iteration
is within a hypersphere with a predefined radius $\rho$, centered
about the best of the generated vectors $\boldsymbol{x}_{1}^{(k)}$.
In both cases, the found best $\boldsymbol{m}^{(k)}$ is declared
as the solution.

The initial value of $\boldsymbol{m}^{(0)}$ and $\boldsymbol{P}^{(0)}$
in HKA are computed as follows. 

\begin{equation}
m_{i}^{(0)}=\frac{max(S_{i})+min(S_{i})}{2}\label{init_mean}
\end{equation}

\begin{equation}
p_{i,i}^{(0)}=\frac{max(S_{i})-min(S_{i})}{6}\label{init_var}
\end{equation}

Here $m_{i}^{(0)}$ is the $i^{th}$ component of the mean vector,
$p_{i,i}^{(0)}$ is the $i^{th}$ diagonal of the variance matrix,
while $max(S_{i})$ and $min(S_{i})$ indicate the maximum and minimum
bound value along the $i^{th}$ dimension of the search space $S$.

The convergence properties and other analysis of HKA can be found
in \cite{[16]} where the authors have shown that HKA performs well
with respect to several test functions and attains better results
than other algorithms most of the time. The authors of HKA in \cite{[16]}
have also shown that the number of function evaluations to reach the
solution for HKA is lower than a few other metaheuristic optimization
algorithms and this method was able to reach better results most of
the time. Also, constrained optimization was demonstrated by the authors
in solving the welded beam design problem and in tuning Proportional
Integral Derivative (PID) controllers.

\section{Partitional clustering using HKA\label{sec:Clustering-using-HKA}}

\subsection{Partitional Clustering\label{subsec:Partitional-Clustering}}

Assume a dataset $\boldsymbol{X}=\{\boldsymbol{x}_{1},\boldsymbol{x}_{2},\ldots,\boldsymbol{x}_{n}\}$,
where $\boldsymbol{x}_{i}$, $1\le i\le n$ is a datum represented
as a vector in $d$-dimensional feature space. The task of clustering
is to partition $\boldsymbol{X}$ into a set of $K$ clusters $\boldsymbol{C}=\{\boldsymbol{c}_{1},\boldsymbol{c}_{2},\ldots,\boldsymbol{c}_{K}\}$
such that cluster $\boldsymbol{c}_{i}$ contains data points similar
to each other and dissimilar to the data of other clusters with respect
to some measure of similarity. In crisp partitional clustering all
clusters should be non-empty $\forall_{1\le j\le K}(\boldsymbol{c}_{j}\ne\phi)$,
and having no common elements $\boldsymbol{c}_{j}\cap\boldsymbol{c}_{m}=\phi$
where $1\le j\le K$, $1\le m\le K$ and $j\ne m$ while $\cup_{j=1}^{K}\boldsymbol{c}_{j}=\boldsymbol{X}$.

The centroids for the set $\boldsymbol{C}$ are $\boldsymbol{Z}=\{\boldsymbol{z}_{1},\boldsymbol{z}_{2},\ldots,\boldsymbol{z}_{k}\}$
where $\boldsymbol{z}_{i}$ is given by Eq. (\ref{eq:centroid})

\begin{equation}
\boldsymbol{z}_{j}=\frac{1}{card(\boldsymbol{c}_{j})}\sum_{\boldsymbol{x}_{i}\in\boldsymbol{c}_{j}}\boldsymbol{x}_{i}\label{eq:centroid}
\end{equation}

where $1\le j\le K$ , and $card(.)$ is the cardinality of the set
$(.)$ .

Several algorithms consider clustering as a global optimization problem,
where one or more objective functions representing the cluster quality
is optimized. Here we have used the sum of squared intra-cluster Euclidean
distances as the objective function to minimize, which is given by

\begin{equation}
J(\boldsymbol{Z})=\sum_{\boldsymbol{z}_{j}\in\boldsymbol{Z}}\sum_{\boldsymbol{x}_{i}\in\boldsymbol{X}}\mu_{ij}\|\boldsymbol{x}_{i}-\boldsymbol{z}_{j}\|^{2}\label{eq:objective}
\end{equation}

Where $\|.\|$ is the magnitude of the vector $(.)$ in Euclidean
space, and $\mu_{ij}$ is the cluster assignment value defined as

\[
\mu_{ij}=\begin{cases}
1 & if\,\boldsymbol{x}_{i}\in\boldsymbol{c}_{j}\\
0 & otherwise
\end{cases}
\]

where $1\le i\le n$ and $1\le j\le K$ .

Therefore the objective is to optimize Eq. (\ref{eq:objective}) with
respect to $\boldsymbol{Z}$, on a fixed dataset $\boldsymbol{X}$,
and find the cluster assignment $\mu_{ij}$ for all $i$ and $j$.

Since we have used the ``idea'' of K-Means algorithm in our HKA-K
approach, a brief description of K-Means algorithm is given below.
\begin{enumerate}
\item The number of clusters $K$ is prespecified. Initially, the set of
cluster centers $\boldsymbol{Z}$ is randomly generated.
\item A data point $\boldsymbol{x}_{i}$ is assigned to $\boldsymbol{c}_{j}$
if $\boldsymbol{z}_{j}$ is closest to $\boldsymbol{x}_{i}$, ie.
$\|\boldsymbol{x}_{i}-\boldsymbol{z}_{j}\|\le\|\boldsymbol{x}_{i}-\boldsymbol{z}_{m}\|$,
where $1\le m\le K$ and $j\ne m$. All datapoints are assigned in
this way.
\item Using Eq. (\ref{eq:centroid}), the cluster center $\boldsymbol{z}_{j}$
is updated with the datapoints assigned to the $j^{th}$ cluster in
Step 2. All the cluster centers are updated in this way. 
\item If a termination criterion is satisfied, then the process is stopped.
Else, the control goes to Step 2 with current cluster centers.
\end{enumerate}
A detailed analysis of K-Means approach can be found in \cite{[34]}.

\subsection{HKA based partitional clustering}

For our purpose we represent the cluster centers as a concatenated
vector. If the $K$ cluster centers are $\boldsymbol{z}_{1},\boldsymbol{z}_{2},\ldots,\boldsymbol{z}_{K}$
in $\mathbb{R}^{d}$, then the concatenated vector will be $\boldsymbol{q}=[\boldsymbol{z}_{1},\boldsymbol{z}_{2},\ldots,\boldsymbol{z}_{K}]$,
where $[\,.\,]$ denotes the concatenation of the comma separated
vectors. Clearly, the dimension of vector $\boldsymbol{q}$ will be
$K\ldotp d$. To compute the objective value for such a concatenated
vector, the individual $\boldsymbol{z}_{i}$s are extracted from $\boldsymbol{q}$
and Eq. (\ref{eq:objective}) is used.

The clustering process progresses as in the HKA algorithm. It starts
with an initial mean $\boldsymbol{m}^{(0)}$ and covariance matrix
$\boldsymbol{P}^{(0)}$ for the Gaussian random number generator.
$\boldsymbol{m}_{best}^{(0)}$ is set to $\boldsymbol{m}^{(0)}$.
In every iteration $k$, the process generates $N$ random vectors
from a Gaussian pdf with mean vector and covariance matrix $\boldsymbol{m}^{(k)}$
and $\boldsymbol{P}^{(k)}$, respectively. Next, it computes $\boldsymbol{\xi}^{(k)}$
using Eq. (\ref{eq:measurement}) and $\boldsymbol{V}^{(k)}$ using
Eq. (\ref{eq:measurement_error}) and then estimates $\hat{\boldsymbol{m}}^{(k)}$
using Eq. (\ref{eq:predicted_mean}) and $\hat{\boldsymbol{P}}^{(k)}$
using Eq. (\ref{eq:posterior_variance}) and (\ref{eq:kalman_gain}).
If the objective value of $\hat{\boldsymbol{m}}^{(k)}$ is better
than $\boldsymbol{m}_{best}^{(k)}$ then it is updated with $\hat{\boldsymbol{m}}^{(k)}$.
For the next iteration, $\boldsymbol{m}^{(k+1)}$ and $\boldsymbol{P}^{(k+1)}$
are computed as in Eq. (\ref{eq:m_plus_one}) and Eq. (\ref{eq:posterior_var})
respectively, and $k$ is incremented. This process is repeated until
the value of $k$ reaches some prefixed number of iterations, or the
best $N_{\xi}$ samples generated in some iteration is within a hypersphere
of a predefined radius $\rho$, centered at the best sample point.
When terminated, $\boldsymbol{m}_{best}^{(k)}$ is taken as the solution.
Here, $\boldsymbol{m}_{best}$, $\hat{\boldsymbol{m}}$, $\boldsymbol{\xi}$
and the generated vectors from the Gaussian pdfs are encoded as described
in the previous paragraph.

For initialization, $\boldsymbol{m}^{(0)}$ is chosen in a similar
way as in HKA but here the search space is considered as the minimum
bounding hyperbox of the dataset.

\begin{equation}
\boldsymbol{m}^{(0)}=cat(\boldsymbol{\bar{x}},K);\,\bar{x}_{j}=\frac{max_{X}(\boldsymbol{x}_{j})+min_{X}(\boldsymbol{x}_{j})}{2}\,\,\,\,;1\le j\le d\label{eq:m_init}
\end{equation}

Here $cat(\boldsymbol{v},K)$ is the concatenation of the vector $\boldsymbol{v}$,
$K$ times. $max_{X}(\boldsymbol{x}_{j})$ (and $min_{X}(\boldsymbol{x}_{j}))$
denotes the maximum (and minimum) value of dimension $j$ over all
datapoints in the dataset.

The initial covariance matrix $\boldsymbol{P}^{(0)}$ is taken as 

\begin{equation}
\boldsymbol{P}^{(0)}=\left[\begin{array}{ccc}
(\sigma_{1}^{(0)})^{2} & \cdots & \mathbf{0}\\
\vdots & \ddots & \vdots\\
\mathbf{0} & \cdots & (\sigma_{(K.d)}^{(0)})^{2}
\end{array}\right]\label{eq:var_covar_init_a}
\end{equation}

\begin{equation}
\boldsymbol{\sigma}^{(0)}=cat(\boldsymbol{\tilde{x}},K);\,\tilde{x}_{j}=\frac{max_{X}(\boldsymbol{x}_{j})-min_{X}(\boldsymbol{x}_{j})}{6}\,\,\,\,;1\le j\le d\label{eq:var_covar_init_b}
\end{equation}

Given a dataset, the minimum bounding hyperbox will contain all the
data points inside the hyperbox. As we are performing a centroid based
clustering task and the cluster centers cannot be outside this bounding
box, the algorithm does not need to search outside this hyperbox.
Therefore we select the starting point of the search at the center
of the minimum bounding hyperbox and select a variance which is sufficiently
large so that the random samples covers almost the entire search space.
Also, we limit the search so that the search does not cross this hyperbox.

\section{HKA-K Clustering\label{sec:KHKA-Clustering}}

Though HKA performs fairly well for clustering data, as shown in Section
\ref{sec:Experiment}, it would be interesting to see if the results
could be improved.  Improvement can be envisioned on two aspects,
namely accuracy of the results and efficiency of the algorithm. One
possible way of achieving both is to exploit global search nature
of the algorithm combined with a local search. We have tried to attain
this by exploiting the simplicity and speed of K-Means approach under
a hybrid framework. A centroid based K-Means like step is used because
of its simplicity. A full K-Means algorithm is not run, instead we
only use one single step of centroid updates with the K-Means approach
at each iteration.

In this work we make the following two modifications: (a) we introduce
a new step after the estimation step of HKA with a view to attain
faster convergence and better accuracy, (b) we incorporate a restart
mechanism which in effect attempts to avoid the search getting trapped
in a local optimum or stagnation. We show that these modifications
are effective in improving convergence and speed compared to HKA based
clustering as well as several other state-of-the-art metaheuristic
based clustering algorithms.

For the above process, we introduce a \emph{time update} step which
updates the estimated $\hat{\boldsymbol{m}}^{(k)}$ through one single
step of weighted K-Means operator. The estimated $\hat{\boldsymbol{P}}^{(k)}$
is also updated accordingly. See \cite{[91],[101]} for time update
in Kalman filtering approach. On the other hand, in HKA the estimated
$\hat{\boldsymbol{m}}^{(k)}$ is directly used in the next iteration
(Eq. (\ref{eq:m_plus_one})). By the term K-Means operator we mean
one single iteration of K-Means which will update the centroids to
move towards the solution, we control this movement through weight.
We also introduce a conditional restart mechanism. The search is restarted
if the $N_{\xi}$ sample points fall in a small region limited by
a constant $\epsilon$.

\begin{algorithm*}[t]
\caption{HKA-K\label{alg:KHKA Algorithm}}

\begin{itemize}
\begin{singlespace}
\item \textbf{Step 1}, Select $N$, $N_{\xi}$, $\alpha$, $w$, $maxiter$
 and $\epsilon$. Initialize mean $\boldsymbol{m}^{(0)}$ using Eq.
(\ref{eq:m_init}), covariance matrix $\boldsymbol{P}^{(0)}$ using
Eq. (\ref{eq:var_covar_init_a}), (\ref{eq:var_covar_init_b}). Set
$\boldsymbol{m}_{best}^{(0)}=\boldsymbol{m}^{(0)}$ and $k=0$.
\item \textbf{Step 2}, Randomly generate $N$ number of $(K\ldotp d)$-dimensional
vectors $\boldsymbol{q}(k)=\{\boldsymbol{q}_{1}^{(k)},\boldsymbol{q}_{2}^{(k)},...,\boldsymbol{q}_{N}^{(k)}\}$
from Gaussian pdf $\mathcal{N}(\boldsymbol{m}^{(k)},P^{(k)})$.
\item \textbf{Step 3}, Compute the measurement point $\boldsymbol{\xi}^{(k)}$
using equation Eq. (\ref{eq:measurement}) and the variance $\boldsymbol{V}^{(k)}$
using Eq. (\ref{eq:measurement_error}).
\item \textbf{Step 4}, Compute estimation of the optimum $\hat{\boldsymbol{m}}^{(k)}$
and variance $\hat{\boldsymbol{P}}^{(k)}$ using Eq. (\ref{eq:predicted_mean}),
(\ref{eq:posterior_variance}) and (\ref{eq:kalman_gain}).
\item \textbf{Step 5}, Apply single step of K-Means on $\hat{\boldsymbol{m}}^{(k)}$
to get $\boldsymbol{m'}^{(k)}$ , use Eq. (\ref{eq:weighting}) to
get $\boldsymbol{m}^{(k+1)}$ and Eq. (\ref{eq:variance_progress}),(\ref{eq:A_compute})
to compute $\hat{\boldsymbol{W}}^{(k+1)}$. Apply slowdown factor
and adjust the time updated variance $\boldsymbol{P}^{(k+1)}$ as
in Eq. (\ref{eq:posterior_var-1}).
\item \textbf{Step 6}, If $J\left(\boldsymbol{m'}^{(k)}\right)<J\left(\boldsymbol{m}_{best}^{(k)}\right)$
then set $\boldsymbol{m}_{best}^{(k+1)}=\boldsymbol{m'}^{(k)}$, else
set $\boldsymbol{m}_{best}^{(k+1)}=\boldsymbol{m}_{best}^{(k)}$
\item \textbf{Step} \textbf{7}, If $\rho<\epsilon$ as in Eq. (\ref{eq:rho}),
then reset $k=0$, Generate $\boldsymbol{m}^{(k)}$ by uniformly selecting
$K$ random data points and concatenating them. Generate $\boldsymbol{P}^{(k)}$
as per Eq. (\ref{eq:var_covar_init_a}), (\ref{eq:var_covar_init_b}). 
\item \textbf{Step 8}, If $k>maxiter$ then terminate the algorithm. Output
$\boldsymbol{m}_{best}^{(k)}$ as the solution. Else set $k=k+1$,
goto Step 2
\end{singlespace}
\end{itemize}
\end{algorithm*}

Algorithm \ref{alg:KHKA Algorithm} describes HKA-K in steps. After
initialization in Step 1, $N$ vectors $\boldsymbol{q}(k)=\{\boldsymbol{q}_{1}^{(k)},\boldsymbol{q}_{2}^{(k)},\ldots,\boldsymbol{q}_{N}^{(k)}\}$
are generated in Step 2 randomly from the Gaussian pdf at the $k^{th}$
iteration using the mean $\boldsymbol{m}^{(k)}$ and covariance matrix
$\boldsymbol{P}^{(k)}$. In this context, the mean and covariance
matrix are the parameters to the Gaussian pdf which is modified by
the Kalman filter framework as in HKA. These parameters should not
to be confused with the mean or covariance matrix of the data set.

Next, in Step 3, the measurement $\boldsymbol{\xi}^{(k)}$ is taken
by finding the mean value of the $N_{\xi}$ points as in Eq. (\ref{eq:measurement})
and the measurement error $\boldsymbol{V}^{(k)}$ by Eq. (\ref{eq:measurement_error}).
Next Eq. (\ref{eq:predicted_mean}), (\ref{eq:kalman_gain}), (\ref{eq:posterior_variance})
are used to get the state update $\hat{\boldsymbol{m}}^{(k)}$ and
the related variance $\hat{\boldsymbol{P}}^{(k)}$ in Step 4. The
time update is performed at Step 5, where at first each of the $K$
cluster centroids in $\hat{\boldsymbol{m}}^{(k)}$ are extracted from
the vector representation and the K-Means operator is used to get
the centers $\boldsymbol{m'}^{(k)}$ are obtained. Now, the centers
for the next iteration is given by

\begin{equation}
\boldsymbol{m}^{(k+1)}=\hat{\boldsymbol{m}}^{(k)}+w(\boldsymbol{m'}^{(k)}-\hat{\boldsymbol{m}}^{(k)})\label{eq:weighting}
\end{equation}

The weight $w$ used to control the influence of K-Means can be a
constant or a function. For our experiments we took $w$ as a constant
number in the interval $[0,1]$. A moderate value of $w$ should be
selected such that the effect of HKA and K-Means are balanced.

From the Kalman filtering algorithm, ignoring the control inputs and
process noise, the state and its related covariance matrix for the
next iteration can be obtained as 

\begin{equation}
\boldsymbol{m}^{(k+1)}=\boldsymbol{A}^{(k)}\hat{\boldsymbol{m}}^{(k)}\label{eq:state_progress}
\end{equation}

\begin{equation}
\hat{\boldsymbol{W}}^{(k+1)}=(\boldsymbol{A}^{(k)})\hat{\boldsymbol{P}}^{(k)}(\boldsymbol{A}^{(k)})^{T}\label{eq:variance_progress}
\end{equation}

Here $\boldsymbol{A}^{(k)}$ is the state transition matrix. The related
covariance matrix is updated as shown in Eq. (\ref{eq:variance_progress}).
We find $\boldsymbol{m}^{(k+1)}$ by Eq. (\ref{eq:weighting}) and
hence to update the covariance matrix we need to find $\boldsymbol{A}^{(k)}$.
We consider $\boldsymbol{A}^{(k)}$ to be a diagonal matrix with the
components given by 

\begin{equation}
a_{i,i}^{(k)}=\frac{m_{i}^{(k+1)}}{\hat{m}_{i}^{(k)}}\,\,\,\,;1\le i\le(K\ldotp d)\label{eq:A_compute}
\end{equation}

Where $a_{i,i}^{(k)}$ is the $i^{th}$ diagonal component of $\boldsymbol{A}^{(k)}$.
Conversely in HKA, $\boldsymbol{A}^{(k)}$ is considered identity
matrix and $\boldsymbol{m}^{(k+1)}$ is equal to $\hat{\boldsymbol{m}}^{(k)}$.
Now $\boldsymbol{A}^{(k)}$ is used to update the variance by Eq.
(\ref{eq:variance_progress}). The mean $\boldsymbol{m}^{(k+1)}$
is used in the next iteration and $\boldsymbol{P}^{(k+1)}$ is computed
from $\boldsymbol{W}^{(k+1)}$ after a slowdown step in Eq. (\ref{eq:posterior_var-1})
is used in the next iteration with the Gaussian pdf.

In each iteration the best vector found so far $\boldsymbol{m}_{best}^{(k)}$
is retained in Step 6, which indicates the optimum value of $\boldsymbol{m}^{(k)}$
found till the current iteration. The initial value $\boldsymbol{m}_{best}^{(0)}$
is set to $\boldsymbol{m}^{(0)}$. If $J(\boldsymbol{m'}^{(k)})<J(\boldsymbol{m}_{best}^{(k)})$
then $\boldsymbol{m}_{best}^{(k+1)}$ is set to $\boldsymbol{m'}^{(k)}$.
Otherwise, $\boldsymbol{m}_{best}^{(k)}$ is retained as $\boldsymbol{m}_{best}^{(k+1)}$.
Note that we use the un-weighted K-Means step to store the best vector,
while employ weighted updating for further search.

The slowdown factor $a$ is computed as done in HKA. Therefore the
variance matrix update after the modification along with slowdown
becomes 

\begin{equation}
\boldsymbol{P}^{(k+1)}=\left(\sqrt{\boldsymbol{P}^{(k)}}+a(\sqrt{\boldsymbol{W}^{(k+1)}}-\sqrt{\boldsymbol{P}^{(k)}})\right)^{2}\label{eq:posterior_var-1}
\end{equation}

The restart stage, Step 7 works as follows. At first, the top $N_{\xi}$
vectors from the measurement step $\{\boldsymbol{q}_{1},\boldsymbol{q}_{2},\ldots,\boldsymbol{q}_{N_{\xi}}\}$
are scaled within the range $[0,1]$ to get $\{\boldsymbol{s}_{1},\boldsymbol{s}_{2},\ldots,\boldsymbol{s}_{N_{\xi}}\}$.
Then the maximum Euclidean distances of the vectors centered around
the \emph{best} point $\boldsymbol{s}_{1}$, is computed. Finally,
the distance is scaled by dividing the number of dimensions $K.d$
to get

\begin{equation}
\rho=\frac{max\{\|\boldsymbol{s}_{1}-\boldsymbol{s}_{i}\|\}}{(K\ldotp d)},\,\,\,2\le i\le N_{\xi}\label{eq:rho}
\end{equation}

If $\rho$ drops below a predefined parameter $\epsilon$, an input
to the algorithm, then the search is restarted. Instead of using the
previously mentioned initialization, to start exploring from a different
location, the restart is done by selecting $K$ number of random datapoints
selected uniformly from the dataset and concatenating them to get
a new vector $\boldsymbol{m}^{(0)}$ and the restarted variance $\boldsymbol{P}^{(0)}$
for the restarted process, as in Eq. (\ref{eq:var_covar_init_a})
and (\ref{eq:var_covar_init_b}).

Step 8 terminates the algorithm and takes $\boldsymbol{m}_{best}^{(k)}$
as a solution, if the iterations $k$ exceeds the $maxiter$. Else,
the control goes to Step 2.

\section{Experiment\label{sec:Experiment}}

\subsection{Used Dataset}

To test the usefulness of the proposed HKA-K and HKA clustering algorithms,
we have performed experiments with two synthetic datasets and a few
well known datasets from the UCI Machine Learning repository \cite{[47]}.
We have also compared the performance of HKA-K with some population
based hybrid algorithms. All the algorithms were implemented with
C++ using Armadillo \cite{[48]} linear algebra library. Table \ref{tab:Datasets}
lists the datasets used in our experiment and provides the number
of points $n$, dimensionality $d$ and the number of clusters present
in the data $K$. 

\begin{table}[h]
\begin{centering}
\caption{Datasets used for testing the clustering methods\label{tab:Datasets}}
\par\end{centering}
\centering{}\resizebox{\textwidth}{!}{%
\begin{tabular}{llllllll}
\hline 
 & {\footnotesize{}Artset1} & {\footnotesize{}Artset2} & {\footnotesize{}Iris } & {\footnotesize{}Wine } & {\footnotesize{}Glass } & {\footnotesize{}CMC } & {\footnotesize{}Cancer }\tabularnewline
\hline 
{\footnotesize{}No. of Data: $n$} & {\footnotesize{}600} & {\footnotesize{}250} & {\footnotesize{}150} & {\footnotesize{}178} & {\footnotesize{}214} & {\footnotesize{}1473} & {\footnotesize{}683}\tabularnewline
{\footnotesize{}Dimensions: $d$} & {\footnotesize{}2} & {\footnotesize{}3} & {\footnotesize{}4} & {\footnotesize{}13} & {\footnotesize{}9} & {\footnotesize{}9} & {\footnotesize{}9}\tabularnewline
{\footnotesize{}Clusters: $K$} & {\footnotesize{}6} & {\footnotesize{}5} & {\footnotesize{}3} & {\footnotesize{}3} & {\footnotesize{}6} & {\footnotesize{}3} & {\footnotesize{}2}\tabularnewline
\hline 
\end{tabular}}
\end{table}

\begin{figure}[H]
\begin{centering}
\subfloat[Artset1\label{fig:Artset1}]{\begin{centering}
\includegraphics[width=8cm,height=8cm]{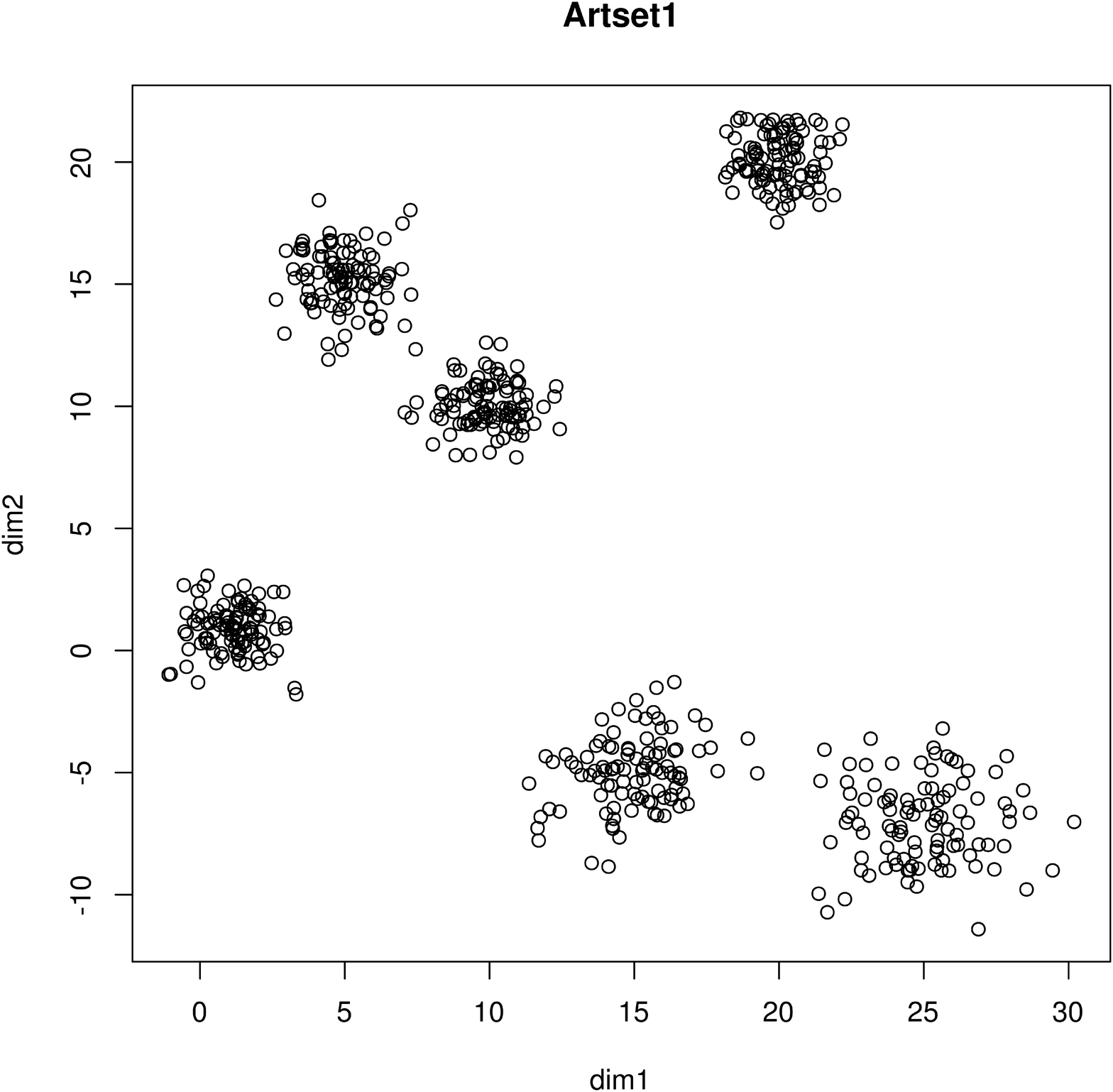}
\par\end{centering}
}
\par\end{centering}
\begin{centering}
\subfloat[Artset2\label{fig:Artset2}]{\begin{centering}
\includegraphics[width=8cm,height=8cm]{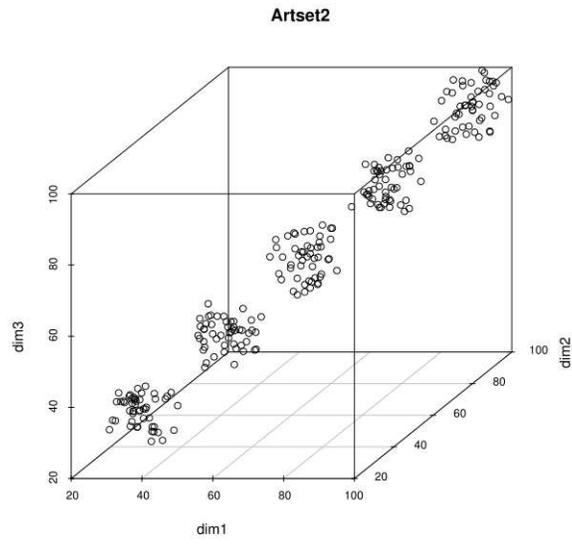}
\par\end{centering}
}
\par\end{centering}
\centering{}\caption{Synthetic Datasets}
\end{figure}

The datasets acquired from the UCI machine learning repository listed
in Table \ref{tab:Datasets} were chosen because these datasets are
quite frequently used in clustering related literature.The synthetic
datasets Artset1 and Artset2 are generated as follows.
\begin{itemize}
\item Artset1 $(n=600,d=2,K=6)$: This two-dimensional dataset consists
of 6 clusters. The plot of this dataset is shown in Figure \figref{Artset1}.
Here 600 patterns were drawn from six independent bi-variate normal
distributions, with the mean and covariances given by \\
$\mathcal{N}\left(\mu=\left(\begin{array}{c}
1\\
1
\end{array}\right),\sum=\left[\begin{array}{cc}
1 & 0\\
0 & 1
\end{array}\right]\right)$, $\mathcal{N}\left(\mu=\left(\begin{array}{c}
5\\
15
\end{array}\right),\sum=\left[\begin{array}{cc}
1.2 & 0\\
0 & 1.2
\end{array}\right]\right)$, $\mathcal{N}\left(\mu=\left(\begin{array}{c}
15\\
-5
\end{array}\right),\sum=\left[\begin{array}{cc}
1.5 & 0\\
0 & 1.5
\end{array}\right]\right)$ , $\mathcal{N}\left(\mu=\left(\begin{array}{c}
10\\
10
\end{array}\right),\sum=\left[\begin{array}{cc}
1 & 0\\
0 & 1
\end{array}\right]\right)$, $\mathcal{N}\left(\mu=\left(\begin{array}{c}
20\\
20
\end{array}\right),\sum=\left[\begin{array}{cc}
1 & 0\\
0 & 1
\end{array}\right]\right)$ , $\mathcal{N}\left(\mu=\left(\begin{array}{c}
25\\
-7
\end{array}\right),\sum=\left[\begin{array}{cc}
2 & 0\\
0 & 2
\end{array}\right]\right)$. A scatterplot for these datapoints are shown in Figure \ref{fig:Artset1}.
\item Artset2 $(n=250,d=3,K=5)$: This three-dimensional synthetic dataset
consists of 5 clusters. The plot of this dataset is shown in Figure
\figref{Artset2}. Each dimension of the dataset is class-wise distributed
as Class 1 - Uniform(85, 100), Class 2 - Uniform(70, 85), Class 3
- Uniform(55, 70) Class 4 - Uniform(40, 55), Class 5 - Uniform(25,
40), where each class has 50 datapoints. A scatterplot for these datapoints
are shown in Figure \ref{fig:Artset2}.
\end{itemize}

\subsection{Used Parameters \label{subsec:Used-Parameters}}

We have tested HKA-K and HKA clustering algorithms, as well as, KGA
\cite{[2]}, GAC \cite{[1]}, ABCC \cite{[32]} and PSO \cite{[21]}
algorithms on the above datasets and compared the results. For all
the algorithms, the optimization criterion used is Eq. (\ref{eq:objective}).

For conducting the experiments, the parameters for HKA clustering
and HKA-K are selected as follows. At first, HKA was run on Glass
dataset with different combinations of $N$, $N_{\xi}$ and $\alpha$,
while HKA-K was run with different combinations of $\alpha$, $N$,
$N_{\xi}$ and $w$. Then the parameter combination with the minimum
sum of intra-cluster distances over 10 executions, (with each execution
having a maximum of 250 iterations) was selected. These parameter
values, given in the first two columns of Table \ref{tab:Parameters-used-for}
have been used for the clustering of all other datasets.

Table \ref{tab:Parameters-used-for} also shows the parameters used
for other four methods with which we compared our algorithm. The Tables
\ref{tab:Results-for-Artset1}-\ref{tab:Results-for-Cancer} show
the results for all the methods with the prescribed parameters listed
in Table \ref{tab:Parameters-used-for}, which are taken from the
corresponding referenced articles. The parameters for KGA and GAC
are taken from \cite{[2]} (though more iterations were added to GAC).
We have used the parameters for ABCC from \cite{[32]}, while the
parameters for the PSO algorithm \cite{[21]} were taken from \cite{[30]}
instead, as we found PSO clustering in \cite{[21]} performed better
with these parameters. All of the KGA, GAC, PSO, ABCC algorithms were
executed 20 times with the corresponding parameters and the results
are shown in Tables \ref{tab:Results-for-Artset1}-\ref{tab:Results-for-Cancer}.
These were used to show how HKA-K and HKA clustering performs when
compared to other algorithms under parameters prescribed by the authors
of the corresponding algorithms. It is very important to note that
the number of function evaluations highly differ to each other for
each algorithms for the given selection of parameters. We have performed
this test to understand the achievable cluster quality of the algorithms.

On the other hand we have performed another experiment with the parameters
mentioned in Table \ref{tab:Parameters-used-equal} which were selected
in such a way that all the algorithms approximately runs equal number
of objective function evaluations. The number of function evaluations
for all the algorithms are kept almost equal to HKA-K in this experiment.
This is done to see how the algorithms converge when all of them evaluate
approximately equal number of functions. Details are explained in
Section \ref{subsec:Benchmark}.

\begin{table}[h]
\caption{Algorithm parameters \label{tab:Parameters-used-for} for results
in Table \ref{tab:Results-for-Artset1}-\ref{tab:Results-for-Cancer}}

\centering{}\resizebox{\textwidth}{!}{%
\begin{tabular}{llllll}
\hline 
{\scriptsize{}HKA-K} & {\scriptsize{}HKA} & {\scriptsize{}KGA} & {\scriptsize{}GAC} & {\scriptsize{}ABCC} & {\scriptsize{}PSO}\tabularnewline
\hline 
{\scriptsize{}$N=20$} & {\scriptsize{}$N=30$} & {\scriptsize{}$pop=50$} & {\scriptsize{}$pop=50$} & {\scriptsize{}$FN=20$} & {\scriptsize{}$pop=10\times K\times d$}\tabularnewline
{\scriptsize{}$N_{\xi}=10$} & {\scriptsize{}$N_{\xi}=6$} &  &  &  & {\scriptsize{}$c1=2.0$}\tabularnewline
{\scriptsize{}$\alpha=0.7$} & {\scriptsize{}$\alpha=0.8$} & {\scriptsize{}$p_{c}=0.8$ } & {\scriptsize{}$p_{c}=0.8$ } & {\scriptsize{}$limit=100$} & {\scriptsize{}$c2=2.0$}\tabularnewline
{\scriptsize{}$\epsilon=0.005$} &  & {\scriptsize{}$p_{m}=0.001$} & {\scriptsize{}$p_{m}=0.001$} &  & {\scriptsize{}$[w_{min},w_{max}]=[0.5,1.0]$}\tabularnewline
{\scriptsize{}$w=0.4$} &  &  &  &  & \tabularnewline
{\scriptsize{}$maxiter=250$} & {\scriptsize{}$maxiter=500$} & {\scriptsize{}$gen=1000$} & {\scriptsize{}$gen=1500$} & {\scriptsize{}$maxiter=1000$} & {\scriptsize{}$maxiter=500$}\tabularnewline
\hline 
\end{tabular}}
\end{table}

In Table \ref{tab:Parameters-used-for} and Table \ref{tab:Parameters-used-equal},
$pop$ is the population size for KGA, GAC and PSO, while $p_{c}$
and $p_{m}$ are the crossover and mutation probability, respectively,
for both KGA and GAC. The parameter $gen$ indicates the number of
generations the algorithms KGA and GAC are allowed to run. The parameters
$c1$ and $c2$ are the cognitive and social factors, and the $w_{min}$
and $w_{max}$ are minimum and maximum momentum factors for PSO. The
parameter $FN$ is the number of food sources to exploit and $limit$
is the limit of the food source for ABCC algorithm. Also, $maxiter$
is the number of maximum iterations taken for HKA-K, HKA clustering,
ABCC and PSO algorithms.

\subsection{Evaluation Metrics}

We have used several metrics to estimate and compare the cluster quality,
which are of two types, namely internal and external. The internal
metrics assume that the datapoints within the clusters are close,
compact and hyper-spherical while different clusters are well separated,
but they do not use the true class labels into consideration. On the
other hand, external metrics employ actual class labels of the dataset.
After the algorithm assigns the data points to the clusters, the quality
is measured using the true class labels of the data. 

We have used Adjusted Rand Index (ARI) \cite{[94]} as an external
validation index. For ARI, higher value indicates better clustering
and the value $1.0$ indicates best result. The ARI is computed as
follows. 

Let $U=\{\boldsymbol{u}_{1},\boldsymbol{u}_{2},\ldots,\boldsymbol{u}_{K}\}$
be the set of clusters found by the algorithm while $V=\{\boldsymbol{v}_{1},\boldsymbol{v}_{2},\ldots,\boldsymbol{v}_{K}\}$
defines the true class label. Let $n$ be the total number of datapoints.
Also, let
\begin{itemize}
\item $a$: Number of pairs of datapoints which are placed in a single cluster
in $U$ and also belongs to a single class in $V$.
\item $b$: Number of pairs of datapoints which are placed in a single cluster
in $U$ but belongs to different classes in $V$.
\item $c$: Number of pairs of datapoints which are placed in different
clusters in $U$ but belongs to a single class in $V$.
\item $d$: Number of pairs of datapoints which are placed in different
clusters in $U$ and also belongs to different classes in $V$.
\end{itemize}
Then the $ARI$ is defined as 

\begin{equation}
ARI=\frac{\binom{n}{2}(a+d)-[(a+b)(a+c)+(c+d)(b+d)]}{\binom{n}{2}^{2}-[(a+b)(a+c)+(c+d)(b+d)]}
\end{equation}

We have also shown the sum of intra-cluster distances (denoted as
``Intra'' in the tables) of the datapoints and Davies-Bouldin index
\cite{[96]} as internal validation metrics. Of them, the sum of intra-cluster
distances (Intra) indicate the compactness of the clusters, denoted
as

\begin{equation}
Intra(\boldsymbol{Z})=\sum_{\boldsymbol{z}_{j}\in\boldsymbol{Z}}\sum_{\boldsymbol{x}_{i}\in\boldsymbol{X}}\mu_{ij}\|\boldsymbol{x}_{i}-\boldsymbol{z}_{j}\|
\end{equation}

In Tables \ref{tab:Results-for-Artset1}-\ref{tab:Results-for-Cancer}
the rows labeled Intra\emph{, }the mean, min, max and std respectively
denote the mean, minimum, maximum and the standard deviation of $Intra(\boldsymbol{Z})$
over the total number of runs for the experiment. Also, the Table
\ref{tab:Results-for-algorithms} states the mean and standard deviations
of the $Intra(Z)$ values.

On the other hand, Davies-Bouldin index finds the ratio of the sum
of within cluster scatter to the between cluster separation. This
metric assumes compact and well separated clusters as ideal. Here,
a lower index value indicates better clustering. Davies-Bouldin index
$DB$ is defined as: 

\begin{center}
$DB=\frac{1}{K}\sum_{i=1}^{K}R_{i}$ where $R_{i}=\max_{i=1,\ldots,K}\{R_{ij}\}\,\,\,\,i=1,\ldots,K;i\ne j$
\par\end{center}

$R_{ij}$ is a similarity measure between clusters $\boldsymbol{c}_{i}$
and $\boldsymbol{c}_{j}$. This is computed based on the measure of
dispersion $S_{i}$ and $S_{j}$ of each of $\boldsymbol{c}_{i}$
and $\boldsymbol{c}_{j}$ respectively, and the dissimilarity between
two clusters $M_{ij}$, which are defined as follows: 

\begin{center}
$R_{ij}=\frac{S_{i}+S_{j}}{M_{ij}}$; $S_{i}=\frac{1}{|\boldsymbol{c}_{i}|}\sum_{x\in\boldsymbol{c}_{i}}\|\boldsymbol{x}-\boldsymbol{z}_{i}\|$;
$M_{ij}=\|\boldsymbol{z}_{i}-\boldsymbol{z}_{j}\|$
\par\end{center}

It may be noted that in Table \ref{tab:Results-for-Artset1}-\ref{tab:Results-for-Cancer}
we have only shown the mean values of ARI and DB index.

The mean execution time in seconds are also shown in Tables \ref{tab:Results-for-Artset1}-\ref{tab:Results-for-Cancer}.
The tests were run in the same system with negligible additional load.
Also, the program executables were generated using the same compiler
and compilation options.

\subsection{Experimental Results\label{subsec:Benchmark}}

The experimental results are presented and evaluated in two ways as
follows. \textbf{(a)} At first, we have presented clustering performance
of the algorithms using the parameters prescribed in the referenced
articles of the other algorithms. We have shown that HKA-K and HKA
clustering can reach similar results in lesser time than the other
methods. \textbf{(b)} Next, we have shown that the performance of
our HKA-K algorithm is similar to KGA and better than other algorithms
when all algorithms are allowed to run for approximately similar number
of function evaluations, with nearly the same execution time.

\subsubsection*{(a) Results with parameters from referenced articles }

The parameters selected for this experiment is mentioned in Table
\ref{tab:Parameters-used-for} and explained in Section \ref{subsec:Used-Parameters}.
The results obtained by other methods as well as HKA-K and HKA clustering
for different datasets are systematically shown in Table \ref{tab:Results-for-Artset1}-\ref{tab:Results-for-Cancer}.
In these tables the best results are shown in bold numerals.

At first, let us compare between HKA and HKA-K methods. From the tables
it is noted that on all datasets the maximum number of iterations
$maxiter$ and the population size $N$ for HKA-K are smaller than
HKA clustering method. Hence HKA-K was able to work faster and attain
similar results with respect to the ARI values for Iris, Wine, Cancer
and CMC dataset, and slightly better for Glass dataset, than the HKA
approach. On the other hand, HKA-K was able to converge to a lower
value of sum of intra cluster distance than HKA clustering.

Next, we compare our results with those of the other methods. To be
unbiased, we have used the same parameters used in the referenced
articles (as given in Table \ref{tab:Parameters-used-for}) for our
simulation. It may be observed from Table \ref{tab:Results-for-Artset1}-\ref{tab:Results-for-Cancer}
that KGA, ABCC and PSO attain a similar result as HKA-K and HKA clustering,
but take more computation time. For the Glass dataset, HKA-K performs
better with respect to the ARI index. On the other hand, GAC was not
able to reach results closer to the other algorithms with the used
parameter settings. With respect to the DB index, we note that except
for the Glass dataset, the values obtained by our methods are the
lowest, which is desirable. For the synthetic datasets, all algorithms
except GAC reached the same values on ARI and DB every time, though
there was slight differences in the mean Intra value.

It is to be noted in Tables \ref{tab:Results-for-Artset1}-\ref{tab:Results-for-Cancer},
that the executions for KGA, GAC, ABCC and PSO need several times
more number of function evaluations compared to HKA-K or HKA clustering.
With the parameters shown in Table \ref{tab:Parameters-used-for},
the approximate number of function evaluations are shown in braces
as follows: KGA (50,000), GAC (75,000), ABCC (40,000), HKA (15,000),
HKA-K (5,250). The number of function evaluations for PSO is different
for different datasets as follows: Artset1 (40,000), for Artset2 (75,000),
Iris (60,000), Wine (195,000), Glass (270,000), CMC (135,000), Cancer
(90,000). The number of function evaluations for PSO vary with respect
to the datasets because the population size depends on the number
of clusters and the number of dimensions of the dataset, as described
in Table \ref{tab:Parameters-used-for}.

The best ARI and DB index are highlighted in the tables. Also, from
Tables \ref{tab:Results-for-Artset1}-\ref{tab:Results-for-Cancer},
for the Intra, DB and ARI index together over all datasets, the number
of times the different algorithms achieved the best results are as
follows: $17$ times by HKA-K, $10$ times by HKA clustering, $17$
times by KGA, $15$ times by ABCC, $7$ times by PSO and $0$ times
by GAC. Therefore, HKA-K and KGA were able to achieve best metric
values maximum number of times over all the metrics and datasets combined.

Though the trend is clear from the mean runtime values, we have performed
a two-tailed Wilcoxon's rank sum test \cite{[99]} for the runtimes.
Wilcoxon's rank sum test is a pairwise test that detects significant
difference between two sample means. The null hypothesis $H_{0}$
for the test states that, \emph{there is no significant difference
between the two samples}. We have taken the significance level as
$\alpha=0.05$ for this two-tailed test. We can reject $H_{0}$ if
the $p$-value for an experiment is less than the selected significance
level and conclude that the difference is significant. The $p$-values
for the Wilcoxon's rank sum test for HKA-K vs each of the other algorithm-dataset
combination had values always less than $10^{-4}$ . We could reject
$H_{0}$ with the mentioned significance level, as the $p$-values
were much below $0.05$. 

\begin{table}[H]
\begin{centering}
\caption{Results for Artset1\label{tab:Results-for-Artset1}}
\par\end{centering}
\centering{}\resizebox{\textwidth}{!}{%
\begin{tabular}{l|l|lll|llll}
\hline 
\multicolumn{1}{l}{Dataset} & \multicolumn{1}{l}{Validity} &  & HKA-K & \multicolumn{1}{l}{HKA} & KGA & GAC & ABCC & PSO\tabularnewline
\hline 
\multirow{7}{*}{Artset1} & \multirow{4}{*}{Intra} & mean & \textbf{918.1220} & 954.2652 & \textbf{918.1220} & 5630.5230 & 918.1235 & 918.1226\tabularnewline
 &  & std & 0.0000 & 142.2113 & 0.0000 & 132.9182 & 0.0150 & 0.0027\tabularnewline
 &  & min & 918.1220 & 918.0910 & 918.1220 & 5358.7400 & 918.1050 & 918.1180\tabularnewline
 &  & max & 918.1220 & 1554.2600 & 918.1220 & 5815.2300 & 918.1550 & 918.1290\tabularnewline
\cline{2-9} 
 & ARI & \multirow{2}{*}{mean} & \textbf{0.9960} & 0.9844 & \textbf{0.9960} & 0.1234 & \textbf{0.9960} & \textbf{0.9960}\tabularnewline
\cline{2-2} 
 & DB &  & \textbf{0.3204} & 0.3347 & \textbf{0.3204} & 7.1917 & \textbf{0.3204} & \textbf{0.3204}\tabularnewline
\cline{2-9} 
 & Time (sec) & mean & \textbf{0.3993} & 1.0327 & 11.5204 & 6.6328 & 2.5441 & 2.7118\tabularnewline
\hline 
\end{tabular}}
\end{table}

\begin{table}[H]
\begin{centering}
\caption{Results for Artset2\label{tab:Results-for-Artset2}}
\par\end{centering}
\centering{}\resizebox{\textwidth}{!}{%
\begin{tabular}{l|l|lll|llll}
\hline 
\multicolumn{1}{l}{Dataset} & \multicolumn{1}{l}{Validity} &  & HKA-K & \multicolumn{1}{l}{HKA} & KGA & GAC & ABCC & PSO\tabularnewline
\hline 
\multirow{7}{*}{Artset2} & \multirow{4}{*}{Intra} & mean & 1782.1000 & 1782.0920 & 1782.1000 & 3906.7310 & \textbf{1782.0770} & 1782.2160\tabularnewline
 &  & std & 0.0000 & 0.0106 & 0.0000 & 220.8353 & 0.0330 & 0.2911\tabularnewline
 &  & min & 1782.1000 & 1782.0700 & 1782.1000 & 3605.4100 & 1781.9900 & 1781.8200\tabularnewline
 &  & max & 1782.1000 & 1782.1100 & 1782.1000 & 4305.4900 & 1782.1000 & 1782.8800\tabularnewline
\cline{2-9} 
 & ARI & \multirow{2}{*}{mean} & \textbf{1.0000} & \textbf{1.0000} & \textbf{1.0000} & 0.3080 & \textbf{1.0000} & \textbf{1.0000}\tabularnewline
\cline{2-2} 
 & DB &  & \textbf{0.5593} & \textbf{0.5593} & \textbf{0.5593} & 3.2681 & \textbf{0.5593} & \textbf{0.5593}\tabularnewline
\cline{2-9} 
 & Time (sec) & mean & \textbf{0.1985} & 0.4537 & 6.8196 & 4.6298 & 1.1767 & 2.7357\tabularnewline
\hline 
\end{tabular}}
\end{table}

\begin{table}[H]
\begin{centering}
\caption{Results for Iris dataset\label{tab:Results-for-Iris}}
\par\end{centering}
\centering{}\resizebox{\textwidth}{!}{%
\begin{tabular}{l|l|lll|llll}
\hline 
\multicolumn{1}{l}{Dataset} & \multicolumn{1}{l}{Validity} &  & HKA-K & HKA & KGA & GAC & ABCC & PSO\tabularnewline
\hline 
\multirow{7}{*}{Iris} & \multirow{4}{*}{Intra} & mean & \textbf{97.3259} & 97.3289 & \textbf{97.3259} & 101.2907 & 97.3262 & 97.3320\tabularnewline
 &  & std & 0.0000 & 0.0096 & 0.0000 & 2.3798 & 0.0012 & 0.0335\tabularnewline
 &  & min & 97.3259 & 97.3129 & 97.3259 & 97.7973 & 97.325 & 97.3258\tabularnewline
 &  & max & 97.3259 & 97.3512 & 97.3259 & 105.1990 & 97.3295 & 97.3471\tabularnewline
\cline{2-9} 
 & ARI & \multirow{2}{*}{mean} & \textbf{0.7302} & \textbf{0.7302} & \textbf{0.7302} & 0.7171 & \textbf{0.7302} & 0.7261\tabularnewline
\cline{2-2} 
 & DB &  & \textbf{0.6623} & \textbf{0.6623} & \textbf{0.6623} & 0.7446 & \textbf{0.6623} & 0.6635\tabularnewline
\cline{2-9} 
 & Time (sec) & mean & \textbf{0.0778} & 0.2281 & 3.2496 & 4.5995 & 0.5436 & 0.6379\tabularnewline
\hline 
\end{tabular}}
\end{table}

\begin{table}[H]
\begin{centering}
\caption{Results for Wine dataset\label{tab:Results-for-Wine}}
\par\end{centering}
\centering{}\resizebox{\textwidth}{!}{%
\begin{tabular}{l|l|lll|llll}
\hline 
\multicolumn{1}{l}{Dataset} & \multicolumn{1}{l}{Validity} &  & HKA-K & \multicolumn{1}{l}{HKA} & KGA & GAC & ABCC & PSO\tabularnewline
\hline 
\multirow{7}{*}{Wine} & \multirow{4}{*}{Intra} & mean & \textbf{16555.7000} & 16574.0800 & \textbf{16555.7000} & 18418.2200 & 16557.4200 & 16557.4100\tabularnewline
 &  & std & 0.0000 & 8.3174 & 0.0000 & 637.6402 & 17.2601 & 0.7564\tabularnewline
 &  & min & 16555.7000 & 16564.2000 & 16555.7000 & 17369.4000 & 16542.3000 & 16556.5000\tabularnewline
 &  & max & 16555.7000 & 16595.0000 & 16555.7000 & 19665.9000 & 16569.1000 & 16558.6000\tabularnewline
\cline{2-9} 
 & ARI & \multirow{2}{*}{mean} & \textbf{0.3711} & \textbf{0.3711} & \textbf{0.3711} & 0.3679 & \textbf{0.3711} & \textbf{0.3711}\tabularnewline
\cline{2-2} 
 & DB &  & \textbf{0.5342} & \textbf{0.5342} & \textbf{0.5342} & 0.6597 & \textbf{0.5342} & \textbf{0.5342}\tabularnewline
\cline{2-9} 
 & Time (sec) & mean & \textbf{0.2010} & 0.578 & 5.1260 & 5.7110 & 1.3319 & 1.6246\tabularnewline
\hline 
\end{tabular}}
\end{table}

\begin{table}[H]
\begin{centering}
\caption{Results for Glass dataset\label{tab:Results-for-Glass}}
\par\end{centering}
\centering{}\resizebox{\textwidth}{!}{%
\begin{tabular}{l|l|lll|llll}
\hline 
\multicolumn{1}{l}{Dataset} & \multicolumn{1}{l}{Validity} &  & HKA-K & \multicolumn{1}{l}{HKA} & KGA & GAC & ABCC & PSO\tabularnewline
\hline 
\multirow{7}{*}{Glass} & \multirow{4}{*}{Intra} & mean & 215.5866 & 216.4745 & \textbf{215.4700} & 288.8954 & 216.1428 & 216.0280\tabularnewline
 &  & std & 0.5211 & 1.3169 & 0.0000 & 13.3002 & 1.4826 & 2.6959\tabularnewline
 &  & min & 213.4210 & 215.7260 & 215.4700 & 273.1690 & 213.7490 & 214.2650\tabularnewline
 &  & max & 215.9210 & 219.7340 & 215.4700 & 318.1390 & 221.4760 & 219.6710\tabularnewline
\cline{2-9} 
 & ARI & \multirow{2}{*}{mean} & \textbf{0.2664} & 0.2645 & 0.2552 & 0.1266 & 0.2583 & 0.2556\tabularnewline
\cline{2-2} 
 & DB &  & 0.9556 & 0.9755 & 0.9351 & 4.7682 & \textbf{0.9092} & 0.9174\tabularnewline
\cline{2-9} 
 & Time (sec) & mean & \textbf{0.3233} & 0.8469 & 7.0956 & 6.4254 & 2.1436 & 16.1743\tabularnewline
\hline 
\end{tabular}}
\end{table}

\begin{table}[H]
\begin{centering}
\caption{Results for CMC dataset\label{tab:Results-for-CMC}}
\par\end{centering}
\centering{}\resizebox{\textwidth}{!}{%
\begin{tabular}{l|l|lll|llll}
\hline 
\multicolumn{1}{l}{Dataset} & \multicolumn{1}{l}{Validity} &  & HKA-K & \multicolumn{1}{l}{HKA} & KGA & GAC & ABCC & PSO\tabularnewline
\hline 
\multirow{7}{*}{CMC} & \multirow{4}{*}{Intra} & mean & 5545.0500 & 5545.2620 & 5545.0500 & 10051.9040 & \textbf{5545.0630} & 5545.7160\tabularnewline
 &  & std & 0.0000 & 0.2267 & 0.0000 & 145.9788 & 0.0747 & 7.1344\tabularnewline
 &  & min & 5545.0500 & 5544.7700 & 5545.0500 & 9729.1200 & 5544.9900 & 5545.2000\tabularnewline
 &  & max & 5545.0500 & 5545.6400 & 5545.0500 & 10212.3000 & 5545.1900 & 5546.3300\tabularnewline
\cline{2-9} 
 & ARI & \multirow{2}{*}{mean} & \textbf{0.0260} & \textbf{0.0260} & \textbf{0.0260} & 0.0052 & \textbf{0.0260} & 0.0259\tabularnewline
\cline{2-2} 
 & DB &  & \textbf{0.7663} & \textbf{0.7663} & \textbf{0.7663} & 4.9677 & 0.7672 & \textbf{0.7663}\tabularnewline
\cline{2-9} 
 & Time (sec) & mean & \textbf{1.172446} & 3.1515 & 21.2861 & 16.7148 & 8.2142 & 38.9426\tabularnewline
\hline 
\end{tabular}}
\end{table}

\begin{table}[H]
\begin{centering}
\caption{Results for Cancer dataset\label{tab:Results-for-Cancer}}
\par\end{centering}
\centering{}\resizebox{\textwidth}{!}{%
\begin{tabular}{l|l|lll|llll}
\hline 
\multicolumn{1}{l}{Dataset} & \multicolumn{1}{l}{Validity} &  & HKA-K & \multicolumn{1}{l}{HKA} & KGA & GAC & ABCC & PSO\tabularnewline
\hline 
\multirow{7}{*}{Cancer} & \multirow{4}{*}{Intra} & mean & \textbf{2988.4300} & 2988.4390 & \textbf{2988.4300} & 3962.1350 & \textbf{2988.4300} & 2988.4540\tabularnewline
 &  & std & 0.0000 & 0.1353 & 0.0000 & 59.1108 & 0.0000 & 0.0479\tabularnewline
 &  & min & 2988.4300 & 2988.1100 & 2988.4300 & 3878.3700 & 2988.4300 & 2988.4000\tabularnewline
 &  & max & 2988.4300 & 2988.6600 & 2988.4300 & 4072.0300 & 2988.4300 & 2988.5300\tabularnewline
\cline{2-9} 
 & ARI & \multirow{2}{*}{mean} & \textbf{0.8465} & \textbf{0.8465} & \textbf{0.8465} & 0.4159 & \textbf{0.8465} & \textbf{0.8465}\tabularnewline
\cline{2-2} 
 & DB &  & \textbf{0.7573} & \textbf{0.7573} & \textbf{0.7573} & 1.2892 & \textbf{0.7573} & \textbf{0.7573}\tabularnewline
\cline{2-9} 
 & Time (sec) & mean & \textbf{0.4228} & 1.1768 & 9.2882 & 8.9919 & 3.1528 & 3.7636\tabularnewline
\hline 
\end{tabular}}
\end{table}

\subsubsection*{(b) Results with equal number of function evaluations}

The parameters selected in the previous experiment for KGA, GAC, ABCC
and PSO from the referenced articles are mentioned in Section \ref{subsec:Used-Parameters}.
The numbers of function evaluations for these selections differ substantially
from each other and are much larger than what HKA-K and HKA clustering
use, as discussed before.

To show how the algorithms converge when they are allowed to run for
approximately equal number of function evaluations, we perform another
experiment by selecting the parameters of the other algorithms in
such a way that it approximately matches the number of function evaluations
for HKA-K, which is 5250 in this case. The parameters selected for
this experiment are mentioned in Table \ref{tab:Parameters-used-equal}
and explained in Section \ref{subsec:Used-Parameters}. Table \ref{tab:Results-for-algorithms}
shows the mean and standard deviation of the $Intra(Z)$ values computed
over 20 runs for each algorithm-dataset combination for parameters
given in Table \ref{tab:Parameters-used-equal}. Note that Table \ref{tab:Parameters-used-equal}
shows the parameters for each algorithm such that each of them performs
around 6000 function evaluations. From Table \ref{tab:Results-for-algorithms}
it can be seen that HKA-K has converged to a better value of $Intra(Z)$
than HKA and GAC, ABCC, PSO algorithms for a similar number of function
evaluations. On the other hand, KGA was able to converge as good as
HKA-K.

\begin{table}[h]
\caption{Algorithm parameters \label{tab:Parameters-used-equal} for results
in Table \ref{tab:Results-for-algorithms}}

\centering{}\resizebox{\textwidth}{!}{%
\begin{tabular}{llllll}
\hline 
HKA-K & HKA & KGA & GAC & ABCC & PSO\tabularnewline
\hline 
$N=20$ & $N=20$ & $pop=12$ & $pop=12$ & $FN=12$ & $pop=12$\tabularnewline
$N_{\xi}=10$ & $N_{\xi}=10$ &  &  &  & $c1=2.0$\tabularnewline
$\alpha=0.7$ & $\alpha=0.8$ & $p_{c}=0.8$  & $p_{c}=0.8$  & $limit=15$ & $c2=2.0$\tabularnewline
$\epsilon=0.005$ &  & $p_{m}=0.001$ & $p_{m}=0.001$ &  & $[w_{min},w_{max}]=[0.5,1.0]$\tabularnewline
$w=0.4$ &  &  &  &  & \tabularnewline
$maxiter=250$ & $maxiter=250$ & $gen=250$ & $gen=250$ & $maxiter=250$ & $maxiter=500$\tabularnewline
\hline 
\end{tabular}}
\end{table}

\begin{table}
\caption{Intra (Z) values for algorithms on datasets for similar numbers of
function evaluations using parameters from Table \ref{tab:Parameters-used-equal}
\label{tab:Results-for-algorithms}}

\centering{}\resizebox{\textwidth}{!}{%
\begin{tabular}{llll|llll}
\hline 
 & \emph{Intra(Z)} & HKA-K & HKA & KGA & GAC & ABCC & PSO\tabularnewline
\hline 
\multirow{2}{*}{Artset1} & mean & \textbf{918.1220 } & 1758.8000  & \textbf{918.1220} & 6983.3830  & 969.9842  & 967.4472 \tabularnewline
 & std & 0.00 & 815.33 &  0.00 &  75.50 &  25.50 & 26.45\tabularnewline
\hline 
\multirow{2}{*}{Artset2} & mean & \textbf{1782.1000 } & 1823.1655  & 1793.1070  & 6751.3695 & 1839.6760  & 1783.3710\tabularnewline
 & std & 0.00 & 153.01 & 37.04  &  176.01 & 39.74 & 1.11\tabularnewline
\hline 
\multirow{2}{*}{Iris} & mean & \textbf{97.3259} & 97.3537  & \textbf{97.3259 } & 226.2000  & 97.6008 & 98.7280 \tabularnewline
 & std & 0.00 & 0.04 &  0.00 & 9.89 &  0.34 & 0.82\tabularnewline
\hline 
\multirow{2}{*}{Wine} & mean & \textbf{16555.7000 } & 17066.4850  & \textbf{16555.7000 } & 36260.9500  & 16688.7450  & 16654.0500\tabularnewline
 & std & 0.00 & 391.70 &  0.00 & 1476.84 & 196.42 &  56.40\tabularnewline
\hline 
\multirow{2}{*}{Glass} & mean & \textbf{215.7021} & 223.9402  & 229.7236  & 397.2842  & 269.1553  & 229.3859\tabularnewline
 & std & 0.85 & 6.13 & 18.05 & 8.71 & 17.24 &  3.91\tabularnewline
\hline 
\multirow{2}{*}{CMC} & mean & 5545.0500  & 5617.8940  & \textbf{5544.9985 } & 11141.6650  & 5570.1785 & 5676.7885 \tabularnewline
 & std & 0.00 & 20.22 & 0.83 & 36.97 &  23.16 & 37.20\tabularnewline
\hline 
\multirow{2}{*}{Cancer} & mean & \textbf{2988.4300 } & 3667.6215  & \textbf{2988.4300 } & 4927.9800  & 2996.6850 & 3137.9700 \tabularnewline
 & std & 0.00 & 201.83 & 0.00 & 33.85 &  13.73 & 0.00\tabularnewline
\hline 
\end{tabular}}
\end{table}

\begin{table}
\caption{HKA-K vs All, Wilcoxon's Rank Sum test p-values, Level of significance
$\alpha=0.025$. For the underlined values $H_{0}$ \emph{could not
}be rejected. For other values $H_{0}$ was rejected. \label{tab:Wilcoxon's-rank-sum}}

\centering{}\resizebox{\textwidth}{!}{%
\begin{tabular}{llllll}
\hline 
HKA-K vs & HKA & KGA & GAC & ABCC & PSO\tabularnewline
\hline 
Artset1 & $1.907\times10^{-6}$ & \uline{$1.0000$} & $9.537\times10^{-7}$ & $9.537\times10^{-7}$ & $9.537\times10^{-7}$\tabularnewline
Artset2 & $1.907\times10^{-6}$ & \uline{$1.0000$} & $9.537\times10^{-7}$ & $9.537\times10^{-7}$ & $9.537\times10^{-7}$\tabularnewline
Iris & $0.003648$ & \uline{$1.0000$} & $9.537\times10^{-7}$ & $0.0002413$ & $9.537\times10^{-7}$\tabularnewline
Wine & $9.537\times10^{-7}$ & \uline{$1.0000$} & $9.537\times10^{-7}$ & $0.00243$ & $9.537\times10^{-7}$\tabularnewline
Glass & $9.537\times10^{-7}$ & $0.0003175$ & $9.537\times10^{-7}$ & $9.537\times10^{-7}$ & $9.537\times10^{-7}$\tabularnewline
CMC & $4.778\times10^{-5}$ & \uline{$0.3051$} & $9.537\times10^{-7}$ & $4.778\times10^{-5}$ & $9.537\times10^{-7}$\tabularnewline
Cancer & $9.537\times10^{-5}$ & \uline{$1.0000$} & $9.537\times10^{-7}$ & \uline{$0.0265$} & $9.537\times10^{-7}$\tabularnewline
\hline 
\end{tabular}}
\end{table}

To understand if HKA-K results were significantly better than the
other algorithms we conduct a one-tailed Wilcoxon's rank sum test
for this experiment. The test is done with HKA-K vs all other algorithms
for each of the datasets with respect to the $Intra(Z)$\emph{ } values
for this experiment. The null hypothesis $H_{0}$ is the same as defined
before. The alternative hypothesis is, \emph{HKA-K has a lower Intra(Z)
value than the same of the compared algorithm}. The level of significance
is kept as $\alpha=0.025$ for this one-tailed test. The results are
shown in Table \ref{tab:Wilcoxon's-rank-sum}. The values in the cells
of the Table \ref{tab:Wilcoxon's-rank-sum} are the $p$-values resulted
from a one-tailed Wilcoxon's rank sum test with HKA-K vs the algorithm
in the corresponding column, for the dataset in the corresponding
row. The underlined values in the table are those which has a $p$-value
\emph{greater} than selected significance level $\alpha=0.025$. We
can reject $H_{0}$ for the experiments for which the $p$-value is
less than $\alpha$, thus indicating the difference is significant.
If the $p$-value is greater than or equal to $\alpha$, then we cannot
make any definite conclusion. Therefore, for the underlined values,
we \emph{cannot} reject $H_{0}$ and make any definite conclusion.

The Tables \ref{tab:Results-for-algorithms} and \ref{tab:Wilcoxon's-rank-sum}
indicate that HKA-K is better than all, except KGA, for which we could
not reject the null hypothesis $H_{0}$ for any datasets except Glass.
For the Cancer dataset, HKA-K had a lower \emph{Intra(Z)} value than
that of ABCC, but the difference was not significant as the p-values
were marginal. In other words, HKA-K is as good as KGA and neither
could decisively outperform each other. For the other algorithms they
reached a worse result in using the same number of function evaluations.
The similar kind of results for HKA-K and KGA is because both of them
uses a K-Means like operator, which guides the solution to a better
point quickly. Although both of them are prone to converge to a local
optimum. KGA avoids it by the mutation operator, while HKA-K avoids
it with the conditional restart step and a carefully selected value
of $w$.

This shows that, given a similar number of function evaluations, the
convergence of HKA-K is significantly better than HKA, GAC, ABCC and
PSO for all the datasets and better than KGA with respect to the Glass
dataset. Although HKA-K does not have a significantly different performance
compared to KGA, indicating both perform nearly the same.

\section{Discussion and Conclusion \label{sec:Conclusion}}

The objective of this work was to demonstrate that clustering can
be performed in the HKA framework and its improvement can be another
state-of-the-art approach in data clustering. By testing on benchmark
datasets we showed that HKA-K can achieve clustering as good as the
compared algorithms with respect to internal and external validation
metrics in lesser time and substantially lesser number of function
evaluations. We have also shown HKA-K was able to converge to a significantly
better value than the other compared algorithms except KGA, when they
all run for similar number of function evaluations. KGA and HKA-K
performances are of similar quality.

There is a basic difference between our method and other clustering
methods we have compared with. Though the approach in this paper is
also a population based method, only one solution per iteration is
generated, which approaches to an optimal point, as mentioned in Section
\ref{sec:HKA}. For the other algorithms each point of the population
represents a possible solution, which moves within the search space,
and is kept in the memory. Whereas for our approach, the population
generated using the Gaussian pdf guided by its mean and the covariance
matrix, which together determines a single solution for an iteration,
which is only kept in the memory. This may lead to a premature convergence
of the algorithm. We have attempted to overcome this by using the
conditional restart stage.

As HKA-K uses one step of K-Means as described in Section \ref{sec:KHKA-Clustering},
the weight for this operator as well as the other parameters has to
be determined in such a way that it does not lead to a premature convergence
or does not spend too much time to search. Although we have set values
for the parameters through a data driven procedure, it may not work
equally well for all kinds of data sets.

In future it would be interesting to investigate other hybridisations
with HKA. A study to improve the performance of HKA as a metaheuristic
population based optimization algorithm can be investigated too. Also,
a new and ingenious multi-solution framework can be devised for this
proposed methodology in place of the conditional restart method.

\bigskip{}

\textbf{Acknowledgements:} We sincerely thank the unknown reviewers
for their constructive comments to improve the quality of the paper.


\bibliographystyle{elsart-num-sort}
\bibliography{hkak_final_sub}

\end{document}